# Context Is What You Need: The Maximum Effective Context Window for Real World Limits of LLMs

Norman Paulsen
Denver, Colorado, USA
Accion Group
npaulsen@acciongroup.com

## Abstract

Large language model (LLM) providers boast big numbers for maximum context window sizes. To test the real-world use of context windows, we 1) define a concept of maximum effective context window, 2) formulate a testing method of a context window's effectiveness over various sizes and problem types, and 3) create a standardized way to compare model efficacy for increasingly larger context window sizes to find the point of failure. We collected hundreds of thousands of data points across several models and found significant differences between the reported Maximum Context Window (MCW) size and the Maximum Effective Context Window (MECW) size. Our findings show that the MECW is not only drastically different from the MCW but also shifts based on the problem type. A few top-of-the-line models in our test group failed with as few as 100 tokens in context; most had severe degradation in accuracy by 1000 tokens in context. All models fell far short of their Maximum Context Window by as much as >99%. Our data reveals the Maximum Effective Context Window shifts based on the type of problem provided, offering clear and actionable insights into how to improve model accuracy and decrease model hallucination rates.

## 1 Introduction

The rise of large language models (LLMs) such as ChatGPT, Claude, Gemini, and LLaMA has reshaped the landscape of natural language processing (NLP), enabling increasingly sophisticated contextual understanding, summarization, coding, and dialog capabilities. Central to the application of these advancements is the concept of the context window (or maximum context length), the maximum number of input tokens (words, punctuation, and symbols) a model can consider at one time. The expansion of context windows from hundreds to tens of thousands and, more recently, to millions of tokens, represents a technical triumph. Several methods have been employed to effectively extend context windows [42], yet an unresolved and often misunderstood question remains: how much of that context can truly be used effectively by the model?

While model specifications cite a maximum context window of 128k, 1 million, or even as much as 10 million tokens [33], these numbers reflect architectural or implementation limits, not necessarily the model's practical capacity for handling or retaining that full input context. Empirical evidence increasingly suggests a divergence between the maximum context window (MCW) and the maximum effective context window (MECW) — the point beyond which additional tokens no longer meaningfully contribute to model output quality (A2.1). Understanding this discrepancy is vital for the efficient deployment of LLMs in domains that demand long-context comprehension, such as legal reasoning, scientific literature synthesis, financial document analysis, and multimodal temporal correlation in video or audio.

This paper explores the emerging distinction between the MCW and the MECW in contemporary transformer-based LLMs. We propose that while LLM architecture permits long-sequence processing, practical limitations constrain the usable span of context in real-world inference tasks. We define MECW as the longest span of token input, for a given problem type, for which incremental tokens degrade the model's output with a measurable effect. This notion reframes the context window not as a flat max input capacity, but as a spectrum of values dependent on the task at hand.

The appeal of longer context windows is intuitive. An LLM capable of digesting entire books, codebases, or sessions without truncation seems closer to achieving general-purpose intelligence. In enterprise settings, longer contexts allow seamless retrieval-augmented generation (RAG), more nuanced chat histories, and document-centric agents capable of reasoning over sprawling datasets. Small context windows limit the practical uses of LLMs.

Yet anecdotal observations from practitioners often contradict this promise. Despite feeding entire books or lengthy transcripts into models with claimed million-token capacity, users frequently observe that LLMs fail to answer questions about information embedded in the input sequences. General observations seem to show performance degrades when prompts rely on large context, and models exhibit increased hallucination rates as token counts rise.

However, there is a lack of empirical evidence to support what we've seen anecdotally in the field. In this paper, we outline a testing methodology for real-world applications of LLMs to find the maximum effect context window. Leveraging the proposed methodology, we've collected hundreds of thousands of data points from the most prominent LLMs on the market. We aggregate that data to show MECW values across a series of different problem types and compare the MCW to the MECW.

This paper makes three key contributions:

- A Formal Definition of MECW: We offer a principled definition of the Maximum Effective Context Window, grounded in informational, theoretical, and behavioral criteria. This includes defining effectiveness in terms of fluctuating measurable influence on model predictions, rather than static inclusion limits like those of MCW.
- Empirical Analysis Across Tasks and Models: We evaluate several state-of-the-art LLMs across a battery of tasks (finding a Needle-in-a-Haystack, finding *Needles*-in-a-Haystack, summarization, finding *Needles*-in-a-Haystack with sorting) using controlled token context intervals to chart the actual usable range of input. This includes both open-source models (e.g., Mistral, LLaMA, Deepseek) and proprietary APIs (e.g., GPT-4o-mini, GPT-5, Claude 3, Gemini 2.5, Gemini 2.0).
- Recommendations for Design and Deployment: Based on our findings, we outline practical guidelines for model architects, prompt engineers, and application developers. These include strategies to optimize RAG pipelines, truncate or summarize distant context, and more realistically context window limit estimates based on MECW rather than MCW.

Broader Implications

Understanding the gap between the maximum context window and maximum effective context windows is not just a technical nuance—it is fundamental to how we effectively use and leverage artificial intelligence in real-world applications. Misinterpretation of context capacity can lead to inefficient system designs, overinvestment in retrieval techniques that yield diminishing returns, or misaligned user expectations. It can also skew benchmarking results, especially when models are assumed to have uniform memory over arbitrarily long sequences.

Moreover, as LLMs are integrated into systems that simulate long-term memory or perform multi-session reasoning, distinguishing between architectural input size limits and real functional capacity becomes crucial. In cognitive science terms, MECW may be more analogous to "working memory" than to "long-term memory," and recognizing that distinction can lead to more robust, interpretable, and grounded model responses.

## 2 Related Work

## 2.1 Tokens Matter

Prior research has shown that long context windows suffer in a few different ways. Existing evidence shows models suffer from a placement of data issue. In older models, providing the same data in different orders produces varying results of success retrieving the requested information. Successful retrieval drops from 76% to 66% by moving the key information from position 1 to position 2, and falls under general model performance (with no data in context) when the key information is moved to position 7. Patterns in attention have been shown to improve data retrieval but are outside the scope of this experiment [19, 31].

Not only does key information placement impact performance, but the size of that information matters. Prior research found that the relevant information token count compared to the total context token count impacts the successful retrieval rate by up to 25% [5].

Notable prior research papers show that models handle context window lengths greater than their training-time sequence length poorly. When using context lengths greater than their training-time sequence length, a U-shaped performance curve emerges based on critical information placement. Additional research shows that models start to degrade at half their training length [2, 31, 43].

Relatively new research posits that model performance on novel tasks, like math and logic problems, suffers from the number of steps needed to complete. The addition of more steps degrades the model's accuracy. We look to show that it's not the number of steps but the token length that causes a breakdown in performance [47].

For the purposes of our experiments, we look to negate data positioning as a contributing factor to performance by randomizing the data for each question proposed to the model. This guarantees an even distribution of data placement throughout the context.

## 2.2 Settings Impacting Performance

Several studies have showcased how model performance varies over a multitude of factors. Model performance relies on several factors, including max allowed output tokens, temperature, top_p, and even the Python frameworks used [18, 48].

Higher temperatures, approaching 1, lead to increased model performance with a tradeoff in reproducibility. For the purposes of our experiments, we use the default temperature of 1. Higher top_p values also lead to improved model accuracy, but without the detriment to stability. We also leave top_p constant at its default values to reduce variability across experiments [18].

Max token values have an outsized impact on long query performance. As models approach set max token limits, they begin to truncate responses and provide unfinished answers. Not only has prior research shown this, but we saw similar results when we started testing while using token limits [12]. As a result, we set all token limits to maximum values to allow models to use as many tokens as necessary to provide an accurate answer.

Reasoning and non-reasoning models work in distinctly different ways, leading to large performance gains from reasoning models. Many experiments have compared the contrasted reasoning vs non-reasoning models of the same provider to help benchmark performance differences between the two [7, 8, 9, 12, 14, 29]. Our research is less interested in the distinction between reasoning and non-reasoning models when it comes to the testing framework. Instead, we focused on the top-performing models from various providers, which were mostly reasoning models.

Fine-tuning models on specific tasks, like data extraction from large documents, increases performance for said tasks. Studies found a 10.5% improvement in data retrieval questions on long context windows by fine-tuning models on synthetic large context window tasks [46]. We did not want to focus the LLM on our tasks and were more interested in model generalization across various tasks.

All settings and frameworks remain constant during our tests to remove as many outside variables as possible. We want to focus on the impact of input context token length as the only variable. To further reduce noise from outside variables, we reran all tests repeatedly until appropriate P-values were achieved (A4).

## 2.3 Novel Question Performance

Standard model performance frameworks are not built to evaluate long context windows. Furthermore, existing evaluation frameworks, like AIME24, AIME25, and GPQA Diamond, all suffer from random seeding volatility, wide fluctuations in scores due to the small number of questions, and variability across different versions of the frameworks [18].

Small datasets, like those used in AIME24, can significantly misrepresent model performance when used in comparisons. The 30 record dataset used in AIME24 means one missed question impacts reported model performance by 3.3%. Many model providers now retest models multiple times on these same small datasets to provide a more accurate number; however, those results are then impacted by seed parameters [18].

The seed parameter, if not explicitly set, is automatically generated dynamically per inference. This was shown to vary model performance on the same dataset significantly higher than the baseline. Coupled with small datasets, this can result in large fluctuations in standardized model performance frameworks.

For all of these reasons, we create a new testing framework for measuring the specific impact of input token length for a given task. None of the existing testing frameworks provides data in a format sufficient for testing incrementally increasing context lengths for real-world use cases.

## 2.4 Similar Frameworks

We did find frameworks with several similarities to our research. These represent the closest research to ours. Our key differentiating factor is a focus on the breaking point of the model/problem combination, emphasizing the point before which the model's accuracy is near perfect.

Models can be extended to effectively find facts in contexts of extraordinary size. The BABILong Benchmark tests model retrieval capacity exceeding 11 million tokens, the equivalent of 16,800 pages or 85 books [24]. This framework focuses on breaking apart long contexts into smaller chunks referenced via recurrent memory. While this is impressive, we are interested in what the optimal-sized chunks are to pass to an LLM.

The LongReason framework provides a set of questions and artificially adds context to the material containing the answer to test models at various context sizes [30]. Per their research, some key limitations are the fixed questions and the non-complexity of the task. It is mostly Needle-in-a-Haystack type problems. Our framework expands on their research by increasing question variability, complexity, and the specificity of token length.

The NoLiMa Benchmark also tests increasing context lengths using Needle-in-a-Haystack style questions, but with a twist. The information requires inference, meaning there is one additional reasoning step required to find the information [36]. They also found similar results to us, that longer contexts degrade performance, but they do not test and compare different problem types across the same model.

The FLenQA data set most closely mirrors our own, but with a few distinctions. FLenQA also focuses on showing model performance degradation over increasingly large context windows [26]. However, they focus on a single type of true/false reasoning question and fill the context with auto-generated text. We diversified the question types and provided only data that could be relevant to the answer, similar to a real-world RAG implementation. We also build upon their findings to show that model degradation on context window size is task-specific, providing guidance in real-world applications.

## 2.5 Other Frameworks

Other frameworks for testing long context windows have been developed in the last 12 months. Several have focused on the Needle-in-a-Haystack problem, demonstrating the effectiveness and limits of finding a single piece of information in a large context window [15, 23, 30, 37]. Others focus on complex tasks on a fixed dataset [6, 10, 22, 27, 50]. None of these focus on incrementally testing model effectiveness on various tasks as the token count increases.

Many additional frameworks test model performance on large contexts for other novel tasks. ETHIC is designed to test long context tasks to see how well LLMs cover the provided material [25]. The DocPuzzle Benchmark provides 100 multi-domain cases with verification mechanisms [50]. CURIE, a scientific long-Context Understanding, Reasoning, and Information Extraction benchmark, also shows models underperform on long contexts [10]. The FACTS Grounding Leaderboard is an ongoing benchmark continuously testing model performance with documents up to 32k tokens in length [22]. Long Code Arena focuses on testing long context windows in a domain-specific benchmark for LLM coding [6].

The LaRA Benchmark also tests large context windows with a focus on RAG vs long-context windows and finds inconclusive results [28]. The tests found many factors are at play, including the model's parameter size, long-text capabilities, context length, task type, and the characteristics of the retrieved chunks. We narrow our focus to context length by task type to determine the relationship.

The HELMET Benchmark tests models with various tasks and context sizes, like our approach [49]. It also finds that models degrade on larger contexts; however they do not focus on finding the point where models degrade for a given task. Instead, they bucket context windows into one of 5 buckets, ranging from 8k to 128k tokens.

Many frameworks prior to 2024 also showed similar limitations with large context windows focusing on a novel set of data and a novel set of questions, like BAMBOO, L-Eval, LongBench, MuLD [1, 4, 13, 20]. Our focus is on building on this great corpus of research by providing practical, real-world guidance

## 3 Methodology

To answer our research questions – how does MECW compare to MCW across models and task complexities, and can this be leveraged to improve model performance – we produced a series of novel questions for LLMs to answer.

This involved creating redundant questions with randomized data, asking these questions of various models repeatedly, slowly increasing the data set size in each round of questions, and recording the answers.

### 3.1 Model Selection

We wanted a wide selection of frontier models from several different providers with open source and proprietary weights. Because of this, we primarily went with reasoning models and excluded small and mid-sized models. Our selection criteria resulted in the following 11 models:

- Open weight: Deepseek.r1-v1:0 [11], Meta.llama3-3-70b-instruct-v1:0 [34]
- Closed weight: claude-3-5-sonnet-20241022 [3], gemini-2.0-flash [16], gemini-2.5-flash-preview-05-20 [17], GPT-4.1 [38], GPT o4-mini [39], GPT-5 [40], Grok-3-latest [45], mistral-medium-2505 [35], Qwen-plus [41]

### 3.1 Framework Design

To collect the necessary data, we developed the following framework to test model performance over an increasing number of input token lengths.

3.1.1 Dataset: We defined our own dataset of 10,000 unique names of individuals. Each individual in the dataset was provided a random number, 1-20, of a random item from a list of 15 possibilities. Each item for each person was then assigned a random color out of 9 possibilities. Example data row:

Abigail Holmes has 19 red balloons.

3.1.2 Question types: We defined four distinct questions based on this dataset. 1) Needle-in-a-Haystack, a search for a single data point from the data set; 2) *Needles*-in-a-Haystack, a search for multiple data points from the data set and then sum the total; 3) Summarization, a full sum of all data points in the data set; 4) Find and sort, a search of multiple data points from the dataset then sorted alphabetically by name. **Exact questions can be found in Appendix A1.

The Needle-in-a-Haystack is the simplest question on our list and, unsurprisingly, the models handled this one the best. For this question, we simply asked for the number of objects a person in the context data had as shown in Figure 1. While all models performed the best at this question type, none managed to effectively find objects up to their MCW.

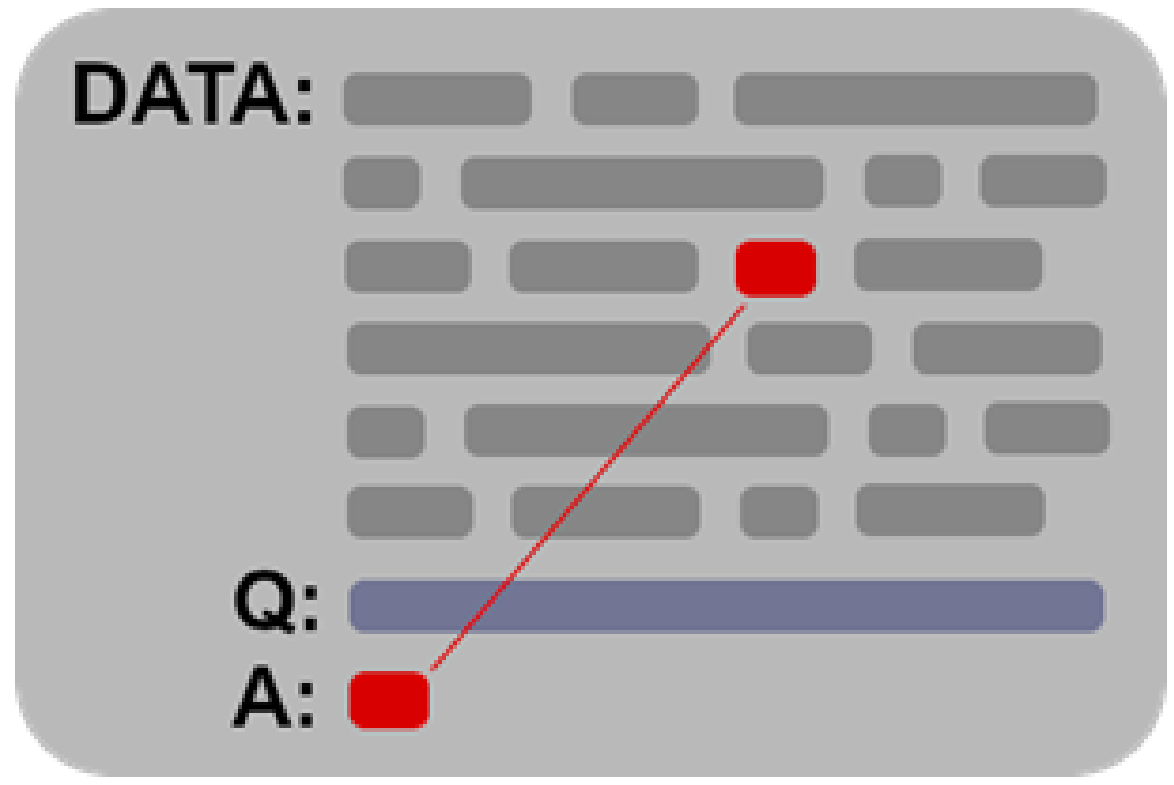

Figure 1: Example of solving the Needle-in-a-Haystack problem by finding the answer in a larger corpus

In the *Needles*-in-a-Haystack question, we asked the model to find all instances of an object type or color (randomly selected) and sum up the total as shown in Figure 2. Here, we saw a large divergence in model performance between the top performers and the lowest performers. The best performers showed reasoning steps that included filtering to the needed information and then summing that shorter list.

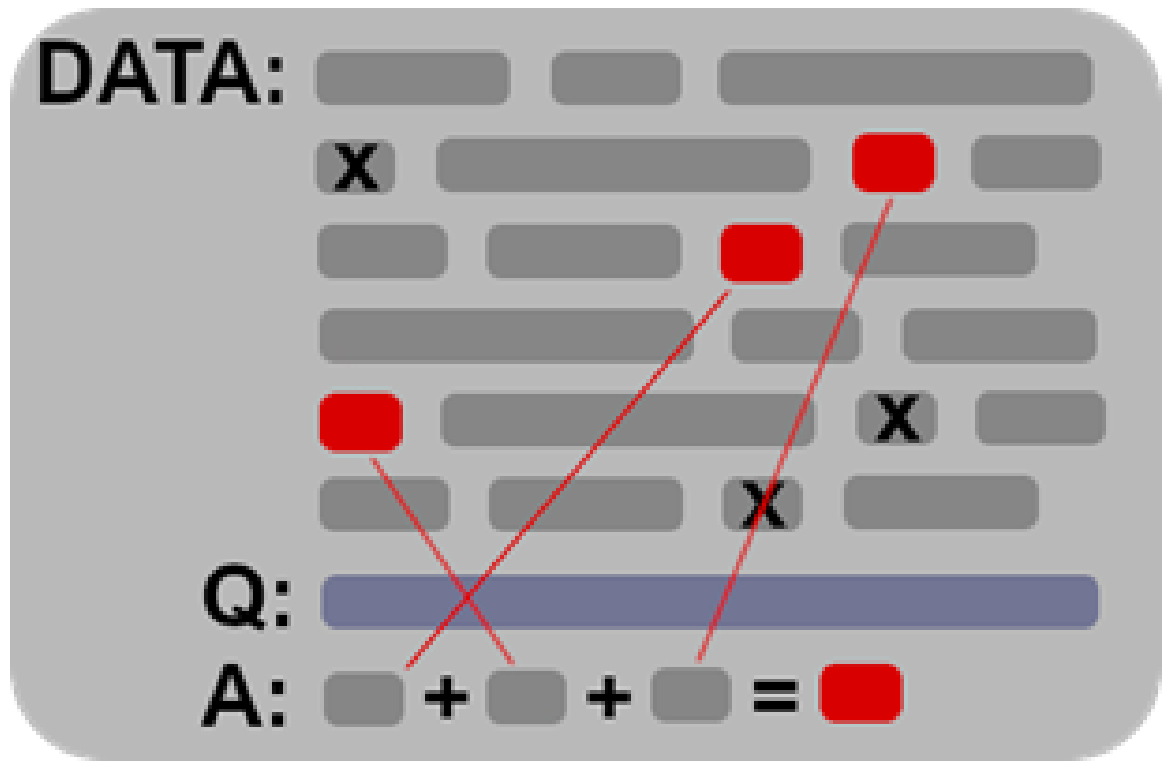

Figure 2: Example of solving the Needles-in-a-Haystack problem by filtering to relevant data in a corpus in any order and summing them into the final answer

For summarization, we simply asked the model to sum up all object totals as shown in Figure 3. All models performed worse than the *Needles*-in-a-Haystack task, which was unexpected. We assumed it would be harder for a model to perform the multi-step problem (filter and sum) over just one step (sum). This further lends itself to the fact that large context windows are ineffective. The filtered list allowed for a smaller context window on the final step, summarization.

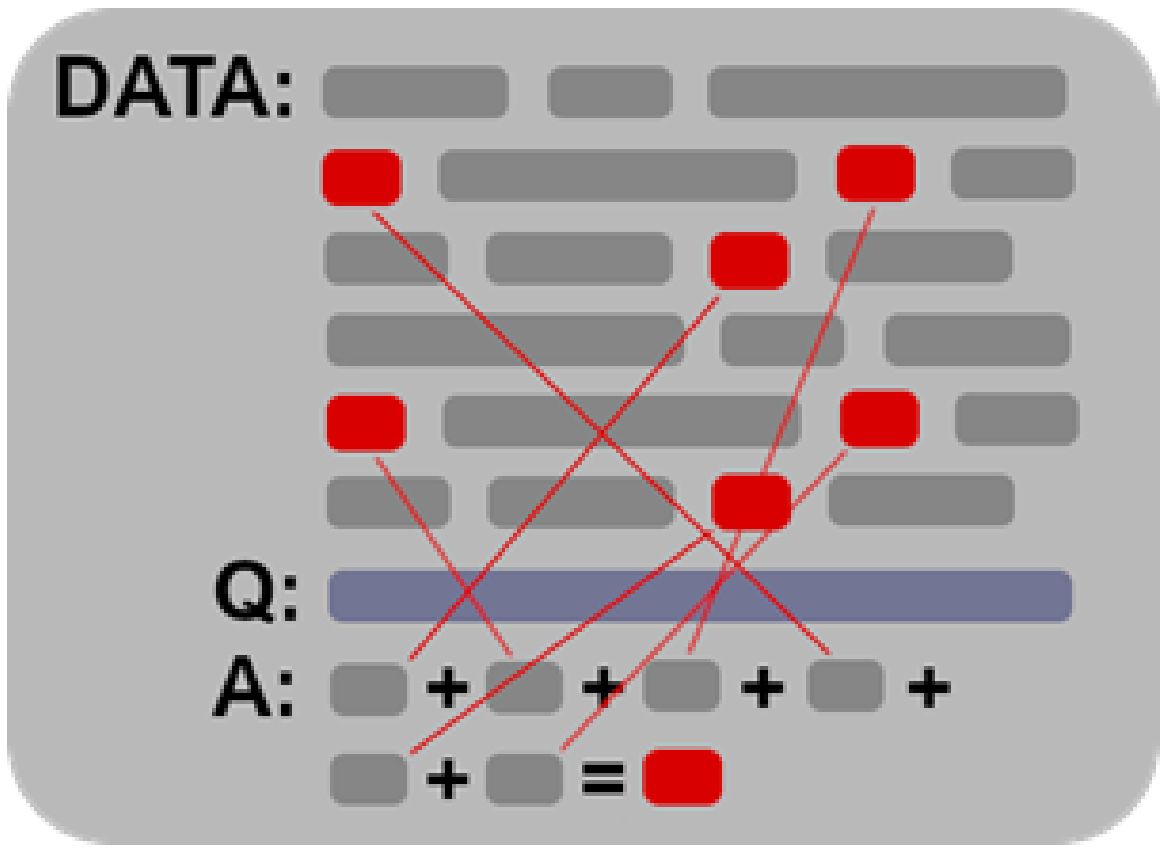

Figure 3: Example of solving the summary problem by extracting all numerical data and summing the values into the final answer

The filter and sort question is the most complex one, requiring a few steps to complete as shown in Figure 4. We ask the model to find the objects of a random type or color, then sort the object counts by owner name, then concatenate the values together in that order.

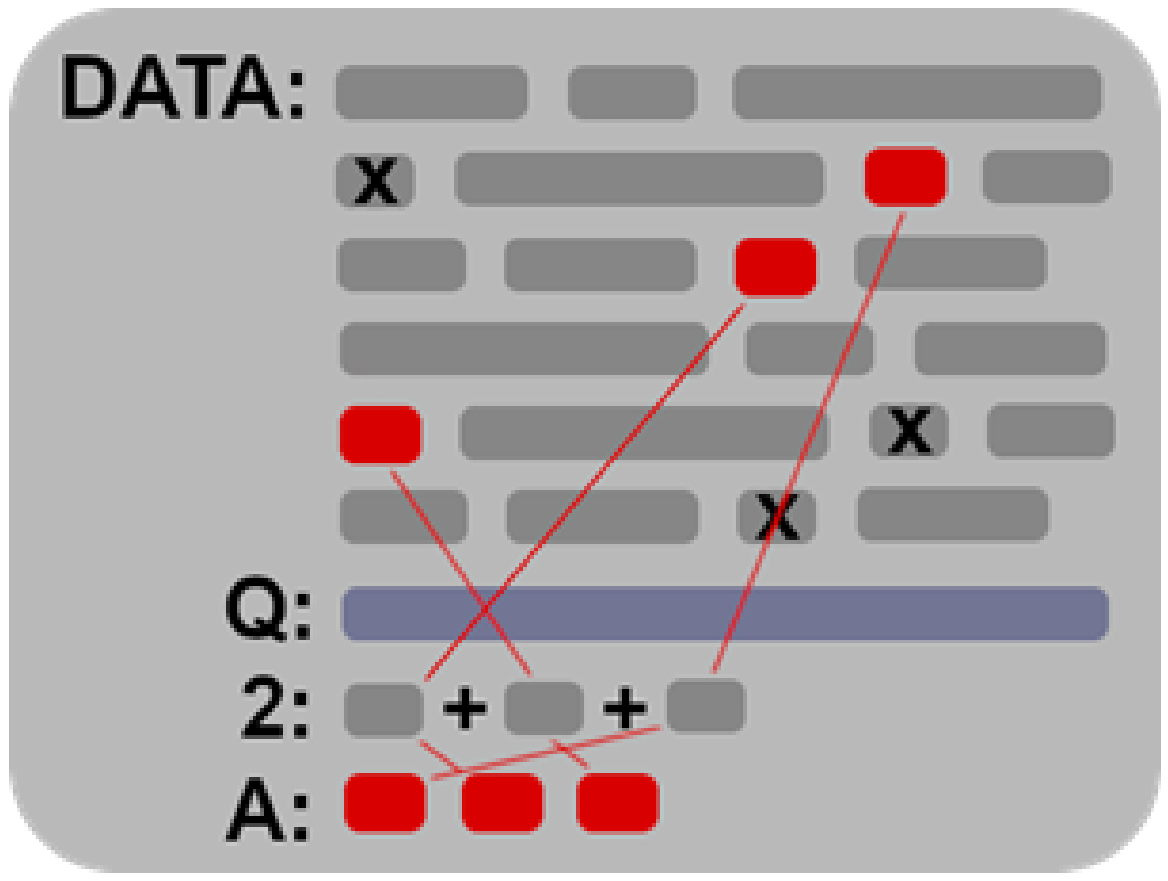

Figure 4: Example of solving the sort problem by filtering to the relevant data and then sorting them into the final answer.

### 3.2 Study Setup

To collect our answers from each model, we connected via API's to every model using Python. We stored our initial dataset and model responses in a Postgres database.

The Python code iterates through a pre-selected range of data points. For each value in that range, we would concatenate that many data points from our data set, formulate a question based on that dataset, and then randomize the order of the dataset. The model instructions were simply to answer the question based on the provided data in a specified JSON format, for ease of retrieval.

The resulting sample dataset and question were then fed into each model selected for a given range of data points. The resulting answer from each model was then captured and compared to the correct answer.

### 3.3 Analysis Procedure

We collected over 66k rows of data, capturing the model name, question type, input token count, and if the correct answer was achieved. For each question and model combination, we validated we captured enough data by measuring the p-value of the relationship between input token count and correct answer (1 for true, 0 for false) as shown in Figures 5-8. Because of the P-values needed for further validation steps (validation of each graphical data point for bucket/model/accuracy combination), the P-values found at this step were always extremely low <1.0e172 (Figures A4).

| Needle-in-a-Haystack | p-value |
|---|---|
| claude-3-5-sonnet-20241022 | 4.05E-244 |
| gemini-2.0-flash | 0.00E+00 |
| gemini-2.5-flash-preview-05-20 | 0.00E+00 |
| gpt-4.1 | 0.00E+00 |
| grok-3-latest | 0.00E+00 |
| mistral-medium-2505 | 4.86E-298 |
| o4-mini | 1.61E-250 |
| qwen-plus | 1.98E-294 |
| us.deepseek.r1-v1:0 | 5.32E-250 |
| us.meta.llama3-3-70b-instruct-v1:0 | 0.00E+00 |

Table 1: P-values of our findings for each model’s answers to the Needle-in-the-haystack question

| *Needles*-in-a-Haystack | p-value |
|---|---|
| claude-3-5-sonnet-20241022 | 0.00E+00 |
| gemini-2.0-flash | 0.00E+00 |
| gemini-2.5-flash-preview-05-20 | 0.00E+00 |
| gpt-4.1 | 0.00E+00 |
| gpt-5 | 0.00E+00 |
| grok-3-latest | 0.00E+00 |
| mistral-medium-2505 | 0.00E+00 |
| o4-mini | 0.00E+00 |
| qwen-plus | 0.00E+00 |
| us.deepseek.r1-v1:0 | 0.00E+00 |
| us.meta.llama3-3-70b-instruct-v1:0 | 0.00E+00 |

Table 2: P-values of our findings for each model’s answers to the *Needles*-in-the-haystack question

| Summary Question | p-value |
|---|---|
| claude-3-5-sonnet-20241022 | 6.97E-183 |
| gemini-2.0-flash | 1.51E-183 |
| gemini-2.5-flash-preview-05-20 | 0.00E+00 |
| gpt-4.1 | 4.44E-191 |
| gpt-5 | 0.00E+00 |
| grok-3-latest | 6.44E-194 |
| mistral-medium-2505 | 8.00E-189 |
| o4-mini | 0.00E+00 |
| qwen-plus | 1.51E-193 |
| us.deepseek.r1-v1:0 | 0.00E+00 |
| us.meta.llama3-3-70b-instruct-v1:0 | 4.05E-206 |

Table 3: P-values of our findings for each model’s answers to the Summarization question

| Find and Sort Question | p-value |
|---|---|
| claude-3-5-sonnet-20241022 | 2.34E-172 |
| gemini-2.0-flash | 2.37E-176 |
| gemini-2.5-flash-preview-05-20 | 0.00E+00 |
| gpt-4.1 | 2.28E-182 |
| gpt-5 | 0.00E+00 |
| grok-3-latest | 1.74E-182 |
| mistral-medium-2505 | 4.40E-178 |
| o4-mini | 0.00E+00 |
| qwen-plus | 2.59E-180 |
| us.deepseek.r1-v1:0 | 0.00E+00 |

| us.meta.llama3-3-70b-instruct-v1:0 | 1.41E-193 |
|---|---|

Table 4: P-values of our findings for each model's answers to the Sorting question

With buckets, we then retested P-values for statistical significance. See Appendix A.3 for P-values for bucketed data.

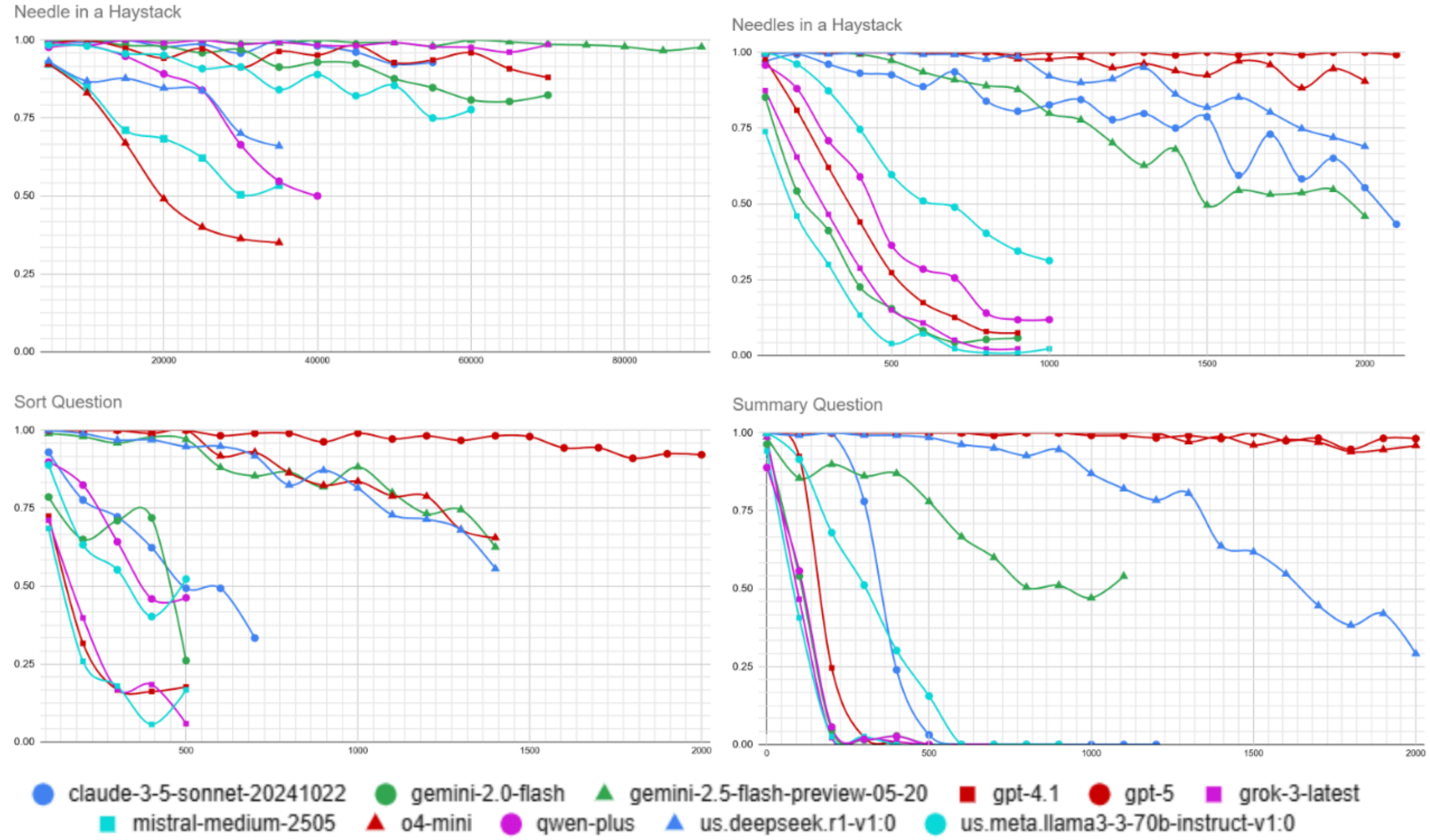

Figure 5: Graphs clearly showing model degradation for each task as token count increases (Y-Axis = Percentage Correct, X-Axis = token count). Full sized graphs can be found in Appendix A3.

To better tie token input count to correct answer rate, we bucketed input token counts into ranges and averaged the correct answers over the range for each model. For the needle in the haystack question, we used buckets of 5000 tokens because of the large level of accuracy across all models for this question type. For the remaining question types, we used buckets of 100 tokens.

To clean the data for bucketing, we removed data points that fell into a bucket with only 1 or 2 data points. This usually occurred on the high end of the token counts, where most tests fell in the preceding buckets, but one rolled into a new bucket. To prevent skewed results (dramatic swings to 0 or 1), we removed these values to eliminate that bucket from the results.

## 4 Findings for Q1: *Does MECW differ from MCW*

Using buckets, clear data patterns emerged. Low levels of token counts improved upon published model hallucination rates with high confidence levels [21]. As the token count increased, all models' accuracy diverged from their published hallucination rates, providing increasingly erratic results as shown in Figure 5. Model performance, in most cases, could be consistently forced to near 0% accuracy levels if provided too large of a context. These findings indicate that there is a need for an MECW measure across models.

## 5 Findings for Q2: *Do different types of questions change the MECW*

Our data provided surprising and clear results in this area. Models perform vastly differently to the type of question asked, as expected. However, we expected model rankings across tasks to remain relatively stable. This was not the case, however. Some models handled the needle in the haystack question far better than their peers, but well underperformed their peers on other question types. This provides an avenue for further research on model performance across task types: coding, scientific research, general Q/A, mathematics, etc.

## 6 Additional Findings

### 6.1 Model Accuracy Using RAG

All models outperformed their standard hallucination rates for our questions, under a certain context size. As context size increased, hallucination rates exceeded base hallucination rates for all models. For the worst-performing models, hallucination rates reached near 100%. The same would likely happen to all models if we continued to test at larger context windows.

Since our line of questioning provided both the facts and the question, it was a simple form of RAG, suggesting, like other research, that RAG increases model accuracy [28]. Our research expands on this to show that accuracy using RAG can reach near 100% levels if utilized under the MECW. It also shows RAG can worsen model performance when exceeding the MECW.

### 6.2 Model Selection

Existing production agentic frameworks tend to utilize the best model or multiple models to guarantee the accuracy of the results. This comes at a detriment to both cost and speed for responses. Understanding the specific use case and MECW for that use case across models allows for a better weighing of cost and speed when making a model selection. While OpenAI o4-mini performed at the top in the needles problem, if we are only utilizing 500 input tokens or less, we could use DeepSeek r1 at a fraction of the price with no reduction in accuracy.

The MECW is designed as an effective way to increase LLM accuracy by measuring, understanding and working within the limits of a given model and problem. This is especially useful for agents in an agentic framework. Each agent is designed with a specific task in mind and the MECW can improve each agent's performance to near-flawless levels. This is without any further modifications, like temperature, top_p, reverse RAG, or a mixture of models to further improve accuracy.

Model rankings also changed across tasks. OpenAI's o4-mini was a top performer in the *Needles* in a Haystack problem but one of the worst at the Needle in a Haystack problem. This further reinforces the need for an MECW measurement to help select the correct model for the correct task.

## 7 Discussion

### 7.1 Implications for GenAI Use

More important than temperature, top_p, seed parameters, and other settings, context window size is the most important factor for determining model accuracy. While these other factors do help model accuracy, they only contribute to overall performance by a few percentage points at most [18, 48]. Context window size can vary a model's performance from near 100% accuracy to near 0% accuracy.

Model Context Windows have grown to outsized amounts as high as 1 million and 10 million tokens. These published limits lead to a false promise of model performance up to that amount. Existing platforms have changed architectures to support these large context windows with the idea that their output performance would improve. Our research shows real-world use cases for LLMs should focus on limiting token count in tasks for best results.

Agentic workflows are the most notable candidate for MECW implementation. As agents are designed to handle a limited scope of tasks, MECW can be applied to increase the accuracy of an agent to nearly 100%. If an agent is failing consistently, splitting the agent into sub-agents with smaller scope and smaller context windows will increase the overall agentic flow's accuracy. This was applied to real-world agentic workflows with promising results at a prior AI startup.

### 7.2 Need for New Testing Frameworks

Existing testing frameworks, like AIME24, AIME25, and GPQA, provide limited value on model performance in real-world use cases. Furthermore, they provide wide swings in measured accuracy because of the small sample rates.

Most applications of Generative AI do not use an LLM alone and leverage some kind of context extension, like RAG. This means we need better testing frameworks for showcasing model performance with more complex use cases. This includes novel question approaches, like those performed by Apple, and context stuffing, like what was performed here [44].

Beyond static testing frameworks, we need a testing framework designed for testing models' MECW across various tasks that can be used by AI developers for their own tasks. Understanding when and where model performance breaks down will help developers understand the limits of a given model in a given context.

## 7.3 Impact on RAG Systems

Our data does support the notion that RAG systems improve hallucination rates. As an example, GPT-5 did not hallucinate once in our dataset when asked a question with under 500 tokens. The problem becomes that as input token amounts increase, the hallucination rate increases. As input token counts reach as few as 2000 tokens, some models' hallucination rates go as high as 99%.

Because of the drastic decline in model performance when using larger context windows, RAG systems leveraging higher token counts decline model performance instead of improving it.

Overall, this leads to cascading failure rates when LLMs are chained together, like in agentic frameworks [32, 47]. The idea of a near-limitless context window leads developers to believe that an agentic system chaining multiple agents with large context windows will perform reasonably well under most situations. As shown through our research, large context windows degrade model performance, so agentic systems relying on large context windows for purposes of RAG will see cascading failures (A2.2).

More importantly, model accuracy can improve above standard hallucination rates simply by providing context windows at the correct size for the model and problem type. This shift in thought prevents cascading model failure by decreasing hallucination rates to a point where chaining agents together will not fail at massively increased rates. Our research concludes that anyone leveraging large context windows and/or RAG systems should be aware of the kinds of questions that they are posing to their models and the limits of context windows around those questions to prevent or reduce hallucination rates and improve overall model accuracy.

## 7.4 Limitations

**Multivariate testing**: Our study focuses on one variable, token count. Isolating this single variable did give us rich results on its impact on LLM performance. However, further testing could be done on token counts tied to other variables, like top-p, to see if another variable allows for larger MECWs.

**Real-world problems**: Our questions and dataset are very simple. Real-world problems might have more structured data input or attached documents, like PDF or excel. Testing the effects of data format could lead to further understanding of how to effectively use a model's context window. Our questions were quite simple. Prompt engineering techniques could be tested to see if there is an uplift in model performance on larger context windows.

## 8 Conclusion

Our findings conclude that the Maximum Context Window does vary widely from the Maximum Effective Context Window (MECW) for all models tested. Additionally, MECW changes with the type of problem presented to the model. While we did not test every model, we hypothesize that these statements hold true for all models currently on the market. Our results suggest that effectively using a model's context window is the highest contributing factor to the hallucination rate of the model.

## Acknowledgements

A special thank you to all of those who inspired this research, pushed me to continue it, and provided support along the way.

# A Appendix

## A1 Survey Questions

For the Needle-in-a-Haystack question, we ask the following:

> How many objects does {person} have?

For the *Needles*-in-a-Haystack, we ask a variant of a question depending on a randomly selected color or object type.

> How many {color} objects are there?

or

> How many {object} are there?

For the summary question, we simply ask:

> How many objects are there total?

For the sorted *Needles*-in-a-Haystack, we ask a variant of a question depending on a randomly selected color or object type.

> Find all people with {color} objects. Sort them by first and last name. Concatenate the number of objects they have into one long string value in the order they were sorted.

or

> Find all people with {object}. Sort them by first and last name. Concatenate the number of objects they have into one long string value in the order they were sorted.

## A2 Definitions

A2.1 Maximum Effective Context Window: The maximum token count, for a given problem type, before the model performance begins to degrade in a measurable fashion.

A2.2 Cascading Failures: where an agentic framework consisting of multiple agents fails most of the time because each agent has a mediocre success rate. A 3-agent system with 70% success per agent results in a system with a 34.3% success rate.

## A3 Graphical Data

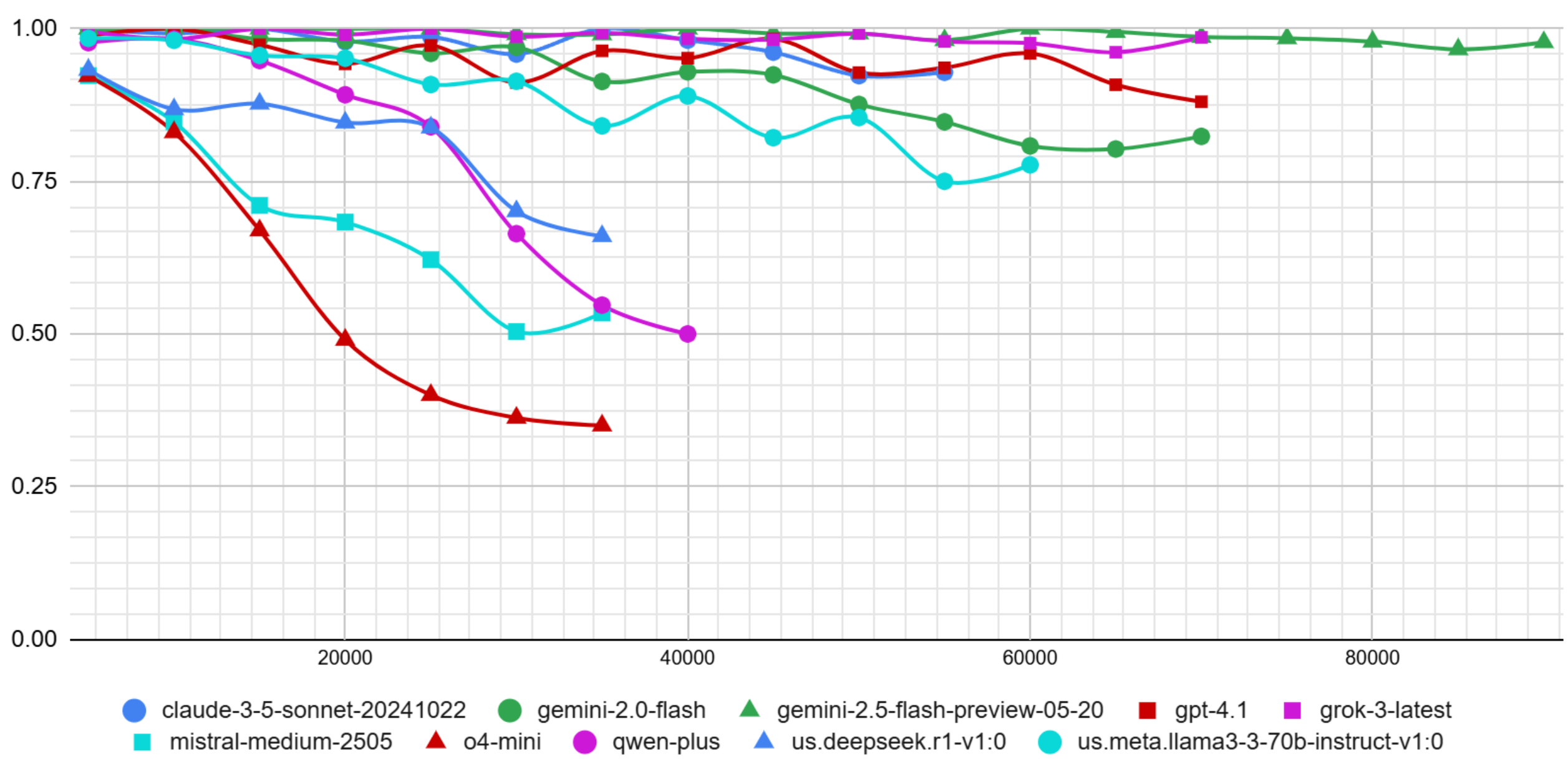

Figure A3.1: Model degradation as token count increases for the Needle-in-a-Haystack question, bucketed by groups of 5000 tokens.

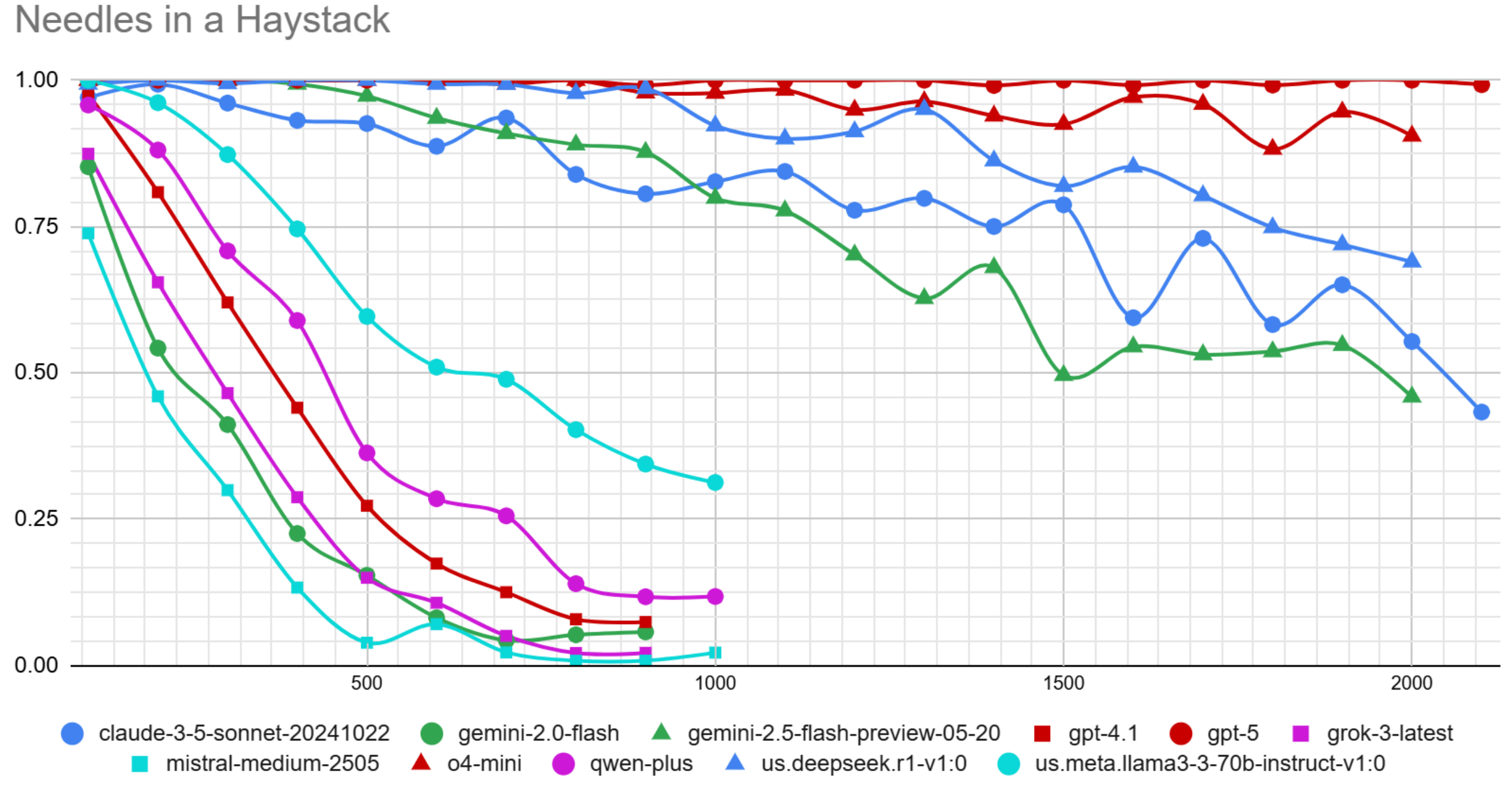

Figure A3.2: Model degradation as token count increases for the *Needles*-in-a-Haystack question, bucketed by groups of 100 tokens.

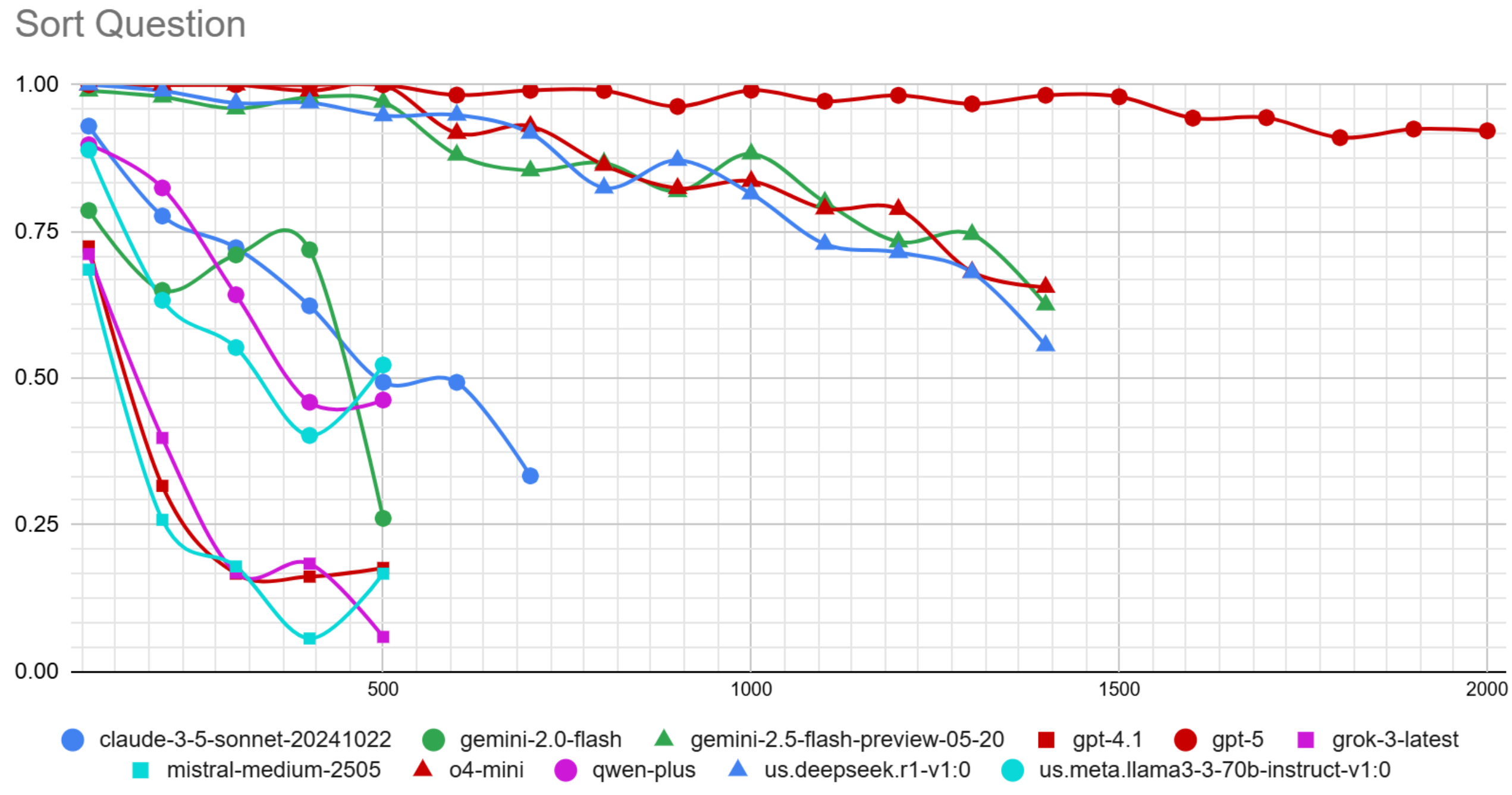

Figure A3.3: Model degradation as token count increases for the Find and Sort question, bucketed by groups of 100 tokens.

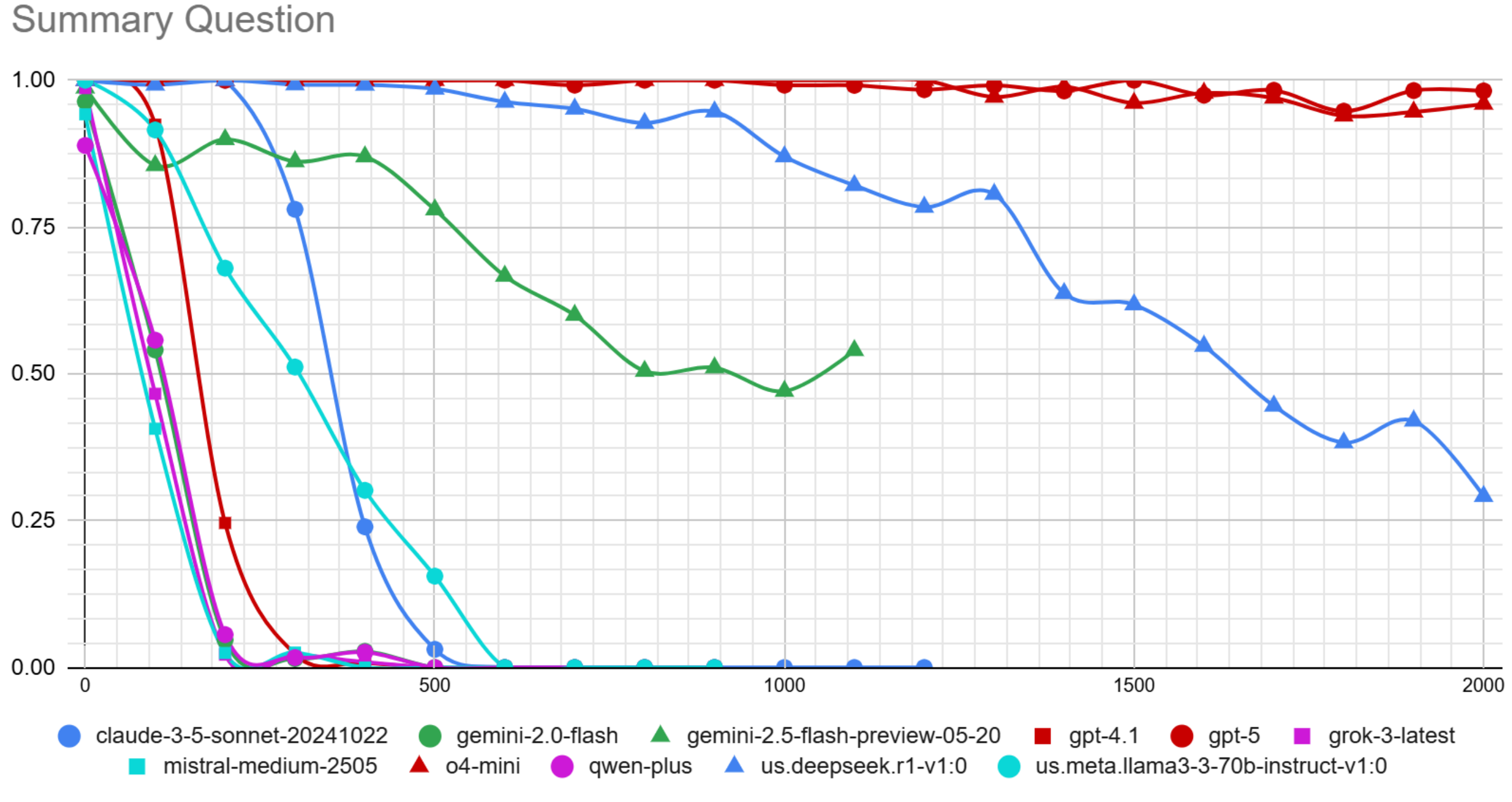

Figure A3.4: Model degradation as token count increases for the Summarization question, bucketed by groups of 100 tokens.

## A.4 P-Value Calculation

Charted P-values for each bucket for each model for each problem set.

### A.4.1 Needle in a Haystack Question

| | claude-3-5-sonnet-20241022 | gemini-2.0-flash | gemini-2.5-flash-preview-05-20 | gpt-4.1 | grok-3-latest | mistral-medium-2505 | o4-mini | qwen-plus | us.deepseek.r1-v1:0 | us.meta.llama3-3-70b-instruct-v1:0 |
|---|---|---|---|---|---|---|---|---|---|---|
| 5000 | 1.01E-164 | 3.14E-184 | 9.35E-186 | 8.69E-185 | 8.96E-182 | 5.90E-188 | 4.51E-183 | 2.18E-188 | 2.55E-186 | 6.76E-187 |
| 10000 | 1.45E-122 | 4.24E-107 | 4.15E-104 | 8.06E-107 | 6.21E-112 | 1.25E-115 | 1.48E-107 | 7.22E-116 | 2.08E-103 | 1.10E-102 |
| 15000 | 9.83E-93 | 6.11E-128 | 5.55E-125 | 7.12E-126 | 4.21E-132 | 1.99E-124 | 3.89E-127 | 1.01E-124 | 1.50E-125 | 1.42E-125 |
| 20000 | 1.11E-111 | 7.24E-125 | 7.18E-127 | 8.92E-126 | 4.64E-123 | 6.57E-149 | 8.33E-126 | 5.98E-149 | 2.33E-126 | 7.66E-126 |
| 25000 | 3.54E-99 | 1.20E-192 | 2.08E-184 | 2.77E-189 | 3.90E-209 | 3.78E-145 | 1.07E-189 | 7.72E-146 | 3.82E-185 | 3.82E-184 |
| 30000 | 8.55E-97 | 1.46E-175 | 6.12E-151 | 6.78E-142 | 1.21E-101 | 1.84E-175 | 6.79E-142 | 3.09E-182 | 1.98E-150 | 4.64E-162 |
| 35000 | 8.76E-152 | 3.68E-149 | 8.49E-164 | 2.84E-169 | 8.17E-194 | 1.07E-195 | 6.22E-159 | 7.57E-152 | 1.68E-159 | 1.60E-172 |
| 40000 | 7.84E-79 | 1.47E-162 | 4.31E-178 | 1.27E-183 | 1.35E-267 | | | 5.38E-26 | | 6.01E-206 |
| 45000 | 9.55E-40 | 5.71E-177 | 6.13E-194 | 4.15E-183 | 2.01E-169 | | | | | 1.25E-166 |
| 50000 | 5.31E-40 | 2.09E-211 | 1.02E-199 | 4.02E-195 | 1.89E-193 | | | | | 2.99E-163 |
| 55000 | 1.05E-92 | 8.98E-211 | 2.11E-257 | 3.56E-255 | 8.64E-229 | | | | | 8.14E-204 |
| 60000 | | 2.45E-249 | 7.64E-238 | 5.29E-243 | 7.87E-279 | | | | | 2.99E-222 |
| 65000 | | 0.00E+00 | 3.42E-303 | 0.00E-02 | 0.00E+00 | | | | | |
| 70000 | | 1.23E-41 | 0.00E+00 | 2.79E-263 | 3.17E-269 | | | | | |
| 75000 | | | 0.00E+00 | | | | | | | |
| 80000 | | | 1.52E-86 | | | | | | | |
| 85000 | | | 2.91E-105 | | | | | | | |
| 90000 | | | 2.57E-86 | | | | | | | |

### A.4.2 *Needles* in a Haystack Question

| | claude-3-5-sonnet-20241022 | gemini-2.0-flash | gemini-2.5-flash-preview-05-20 | gpt-4.1 | gpt-5 | grok-3-latest | mistral-medium-2505 | o4-mini | qwen-plus | us.deepseek.r1-v1:0 | us.meta.llama3-3-70b-instruct-v1:0 |
|---|---|---|---|---|---|---|---|---|---|---|---|
| 100 | 2.65E-75 | 2.59E-136 | 1.07E-113 | 4.53E-109 | 7.64E-81 | 3.73E-136 | 9.31E-118 | 8.10E-131 | 9.29E-113 | 1.50E-129 | 8.53E-105 |
| 200 | 3.30E-137 | 1.07E-177 | 7.62E-150 | 2.15E-153 | 2.68E-105 | 2.55E-181 | 2.37E-163 | 2.25E-174 | 3.98E-165 | 5.22E-174 | 3.93E-173 |
| 300 | 2.54E-143 | 2.35E-194 | 1.58E-173 | 4.66E-172 | 4.12E-119 | 1.37E-202 | 1.52E-182 | 1.36E-193 | 4.03E-183 | 1.93E-191 | 5.55E-197 |
| 400 | 4.98E-156 | 2.70E-216 | 1.39E-188 | 6.51E-192 | 3.02E-136 | 1.07E-217 | 8.73E-207 | 5.31E-216 | 4.48E-202 | 1.58E-205 | 5.99E-217 |
| 500 | 7.09E-176 | 1.23E-203 | 8.32E-192 | 8.38E-199 | 2.55E-141 | 7.25E-215 | 1.32E-197 | 9.69E-210 | 4.32E-198 | 2.60E-216 | 4.03E-206 |
| 600 | 3.56E-166 | 4.58E-213 | 9.18E-191 | 3.97E-192 | 8.98E-148 | 8.60E-214 | 5.73E-196 | 8.16E-204 | 3.53E-199 | 7.57E-208 | 1.79E-212 |
| 700 | 2.76E-174 | 8.17E-198 | 8.19E-185 | 6.95E-192 | 4.83E-159 | 4.21E-200 | 1.87E-191 | 6.00E-211 | 1.74E-192 | 7.12E-205 | 1.06E-192 |
| 800 | 9.50E-173 | 4.07E-200 | 8.98E-189 | 4.66E-188 | 4.76E-159 | 2.77E-212 | 1.55E-189 | 2.10E-207 | 1.38E-191 | 6.51E-200 | 3.83E-209 |
| 900 | 1.24E-157 | 2.74E-119 | 1.10E-185 | 5.94E-132 | 5.95E-176 | 1.40E-87 | 4.03E-190 | 6.69E-216 | 4.63E-195 | 1.71E-216 | 7.06E-189 |
| 1000 | 9.19E-156 | | 2.51E-217 | | 4.17E-173 | | 4.38E-88 | 1.67E-286 | 1.42E-97 | 2.16E-281 | 2.02E-37 |
| 1100 | 5.62E-176 | | 2.22E-221 | | 1.59E-162 | | | 4.35E-287 | | 8.32E-289 | |
| 1200 | 1.03E-146 | | 1.68E-206 | | 1.97E-182 | | | 3.24E-292 | | 4.40E-283 | |
| 1300 | 1.76E-173 | | 2.73E-196 | | 1.01E-193 | | | 8.20E-275 | | 8.42E-271 | |
| 1400 | 1.66E-171 | | 7.71E-255 | | 1.35E-190 | | | 0.00E+00 | | 0.00E-02 | |
| 1500 | 3.61E-186 | | 1.43E-196 | | 6.49E-206 | | | 2.84E-228 | | 7.45E-257 | |
| 1600 | 1.30E-178 | | 6.71E-178 | | 1.94E-210 | | | 8.51E-243 | | 1.79E-238 | |
| 1700 | 9.83E-198 | | 8.56E-260 | | 1.30E-199 | | | 6.54E-263 | | 2.68E-254 | |
| 1800 | 1.24E-163 | | 1.14E-251 | | 9.23E-217 | | | 1.44E-274 | | 2.45E-260 | |
| 1900 | 6.88E-226 | | 2.36E-216 | | 3.48E-224 | | | 1.08E-271 | | 3.37E-223 | |
| 2000 | 3.73E-191 | | 1.74E-160 | | 1.33E-242 | | | 3.36E-203 | | 1.61E-161 | |
| 2100 | 4.75E-126 | | | | 1.65E-246 | | | | | | |
| 2200 | 2.99E-136 | | | | 7.34E-221 | | | | | | |
| 2300 | 2.15E-193 | | | | 6.87E-239 | | | | | | |
| 2400 | 4.02E-187 | | | | 6.13E-207 | | | | | | |
| 2500 | 1.46E-222 | | | | 1.58E-202 | | | | | | |
| 2600 | 3.09E-196 | | | | 1.06E-206 | | | | | | |
| 2700 | 1.58E-190 | | | | 1.14E-235 | | | | | | |
| 2800 | 8.00E-34 | | | | 7.73E-211 | | | | | | |
| 2900 | | | | | 4.46E-218 | | | | | | |
| 3000 | | | | | 1.16E-226 | | | | | | |
| 3100 | | | | | 1.57E-284 | | | | | | |
| 3200 | | | | | 5.74E-229 | | | | | | |

A.4.3 Summarization Question

| | claude-3-5-sonnet-20241022 | gemini-2.0-flash | gemini-2.5-flash-preview-05-20 | gpt-4.1 | gpt-5 | grok-3-latest | mistral-medium-2505 | o4-mini | qwen-plus | us.deepseek.r1-v1:0 | us.meta.llama3-3-70b-instruct-v1:0 |
|---|---|---|---|---|---|---|---|---|---|---|---|
| 0 | 1.19E-38 | 1.03E-50 | 4.92E-50 | 2.99E-50 | 9.51E-13 | 5.19E-53 | 7.28E-48 | 7.39E-50 | 1.43E-47 | 1.31E-50 | 1.52E-40 |
| 100 | 1.45E-72 | 2.53E-97 | 3.97E-95 | 1.59E-97 | 1.38E-91 | 2.85E-99 | 5.22E-91 | 1.54E-97 | 1.48E-90 | 6.32E-95 | 9.40E-96 |
| 200 | 6.20E-92 | 2.96E-122 | 2.89E-122 | 3.59E-124 | 1.53E-113 | 2.66E-129 | 1.59E-115 | 3.91E-125 | 2.74E-118 | 2.24E-122 | 8.52E-120 |

| 300 | 6.10E-100 | 6.91E-145 | 1.55E-142 | 1.54E-143 | 2.27E-127 | 6.50E-149 | 2.48E-131 | 1.04E-143 | 3.53E-131 | 1.78E-144 | 1.22E-139 |
|---|---|---|---|---|---|---|---|---|---|---|---|
| 400 | 8.38E-117 | 1.05E-136 | 1.59E-148 | 2.64E-147 | 2.96E-144 | 5.87E-138 | 2.04E-144 | 7.40E-152 | 1.61E-139 | 6.30E-152 | 1.22E-150 |
| 500 | 3.59E-124 | 6.05E-14 | 9.97E-153 | 6.19E-19 | 4.73E-152 | 1.69E-12 | 7.33E-58 | 2.18E-177 | 1.27E-69 | 6.52E-174 | 2.53E-62 |
| 600 | 2.19E-89 | 1.22E-17 | 1.38E-160 | 1.30E-14 | 4.21E-157 | 8.79E-19 | 8.32E-14 | 1.56E-186 | 1.67E-14 | 1.51E-183 | 1.73E-13 |
| 700 | 8.96E-12 | 1.26E-15 | 1.78E-86 | 1.00E-15 | 3.77E-170 | 1.17E-14 | 3.15E-15 | 5.45E-178 | 5.39E-15 | 2.73E-177 | 4.49E-16 |
| 800 | 3.66E-11 | 3.57E-15 | 9.49E-155 | 2.71E-16 | 3.91E-177 | 8.08E-17 | 5.13E-15 | 5.02E-193 | 1.32E-13 | 4.02E-198 | 4.25E-17 |
| 900 | 5.88E-14 | 9.68E-09 | 1.28E-143 | 8.59E-12 | 3.41E-186 |  | 6.83E-17 | 2.33E-177 | 5.40E-17 | 4.57E-175 | 8.49E-16 |
| 1000 | 1.09E-13 |  | 1.07E-134 |  | 3.03E-188 |  |  | 2.11E-159 |  | 1.36E-157 |  |
| 1100 | 1.34E-09 |  | 1.59E-90 |  | 1.22E-191 |  |  | 1.00E-154 |  | 1.79E-152 |  |
| 1200 | 4.09E-13 |  |  |  | 2.57E-204 |  |  | 7.58E-156 |  | 6.88E-149 |  |
| 1300 |  |  |  |  | 7.20E-194 |  |  | 1.39E-172 |  | 2.05E-164 |  |
| 1400 |  |  |  |  | 1.34E-191 |  |  | 1.20E-156 |  | 6.00E-172 |  |
| 1500 |  |  |  |  | 1.01E-193 |  |  | 2.83E-177 |  | 2.29E-159 |  |
| 1600 |  |  |  |  | 1.22E-207 |  |  | 4.26E-161 |  | 3.94E-167 |  |
| 1700 |  |  |  |  | 7.56E-209 |  |  | 1.69E-175 |  | 1.90E-181 |  |
| 1800 |  |  |  |  | 4.72E-206 |  |  | 3.46E-177 |  | 1.60E-169 |  |
| 1900 |  |  |  |  | 5.93E-206 |  |  | 3.92E-169 |  | 6.20E-182 |  |
| 2000 |  |  |  |  | 6.43E-202 |  |  | 1.82E-183 |  | 3.50E-178 |  |
| 2100 |  |  |  |  | 2.28E-198 |  |  | 2.71E-186 |  | 3.55E-171 |  |
| 2200 |  |  |  |  | 4.38E-215 |  |  | 7.35E-174 |  | 6.49E-188 |  |
| 2300 |  |  |  |  | 8.58E-202 |  |  | 1.50E-189 |  | 1.89E-159 |  |
| 2400 |  |  |  |  | 5.76E-227 |  |  | 7.44E-193 |  |  |  |
| 2500 |  |  |  |  | 4.46E-230 |  |  | 8.22E-188 |  |  |  |
| 2600 |  |  |  |  | 5.61E-226 |  |  | 5.31E-188 |  |  |  |
| 2700 |  |  |  |  | 6.11E-230 |  |  | 1.05E-193 |  |  |  |
| 2800 |  |  |  |  | 8.15E-201 |  |  | 2.71E-190 |  |  |  |
| 2900 |  |  |  |  | 4.43E-231 |  |  | 1.10E-193 |  |  |  |
| 3000 |  |  |  |  | 9.38E-208 |  |  | 1.06E-182 |  |  |  |
| 3100 |  |  |  |  | 5.29E-261 |  |  | 6.74E-222 |  |  |  |
| 3200 |  |  |  |  | 5.74E-284 |  |  | 3.02E-147 |  |  |  |
| 3300 |  |  |  |  | 2.24E-171 |  |  | 1.00E-134 |  |  |  |
| 3400 |  |  |  |  |  |  |  | 3.13E-38 |  |  |  |
| 3500 |  |  |  |  |  |  |  | 2.12E-39 |  |  |  |
| 3600 |  |  |  |  |  |  |  | 1.64E-45 |  |  |  |

A.4.4 Find and Sort Question

| Sorted Question | claude-3-5-sonnet-20241022 | gemini-2.0-flash | gemini-2.5-flash-preview-05-20 | gpt-4.1 | gpt-5 | grok-3-latest | mistral-medium-2505 | o4-mini | qwen-plus | us.deepseek.r1-v1:0 | us.meta.llama3-3-70b-instruct-v1:0 |
|---|---|---|---|---|---|---|---|---|---|---|---|
| 100 | 5.60E-52 | 3.49E-72 | 2.01E-70 | 6.31E-72 | 2.94E-61 | 3.94E-75 | 6.77E-67 | 7.22E-72 | 7.22E-66 | 1.41E-70 | 2.06E-68 |
| 200 | 1.97E-71 | 1.73E-91 | 3.17E-91 | 7.91E-93 | 4.85E-101 | 5.16E-94 | 1.10E-85 | 9.84E-93 | 3.99E-86 | 4.11E-92 | 2.68E-92 |
| 300 | 7.01E-79 | 2.88E-109 | 8.87E-108 | 5.38E-107 | 6.88E-120 | 6.22E-116 | 2.69E-104 | 6.66E-107 | 2.89E-103 | 1.23E-105 | 5.01E-105 |
| 400 | 1.65E-85 | 3.06E-115 | 5.04E-113 | 5.47E-118 | 9.61E-125 | 1.23E-118 | 2.97E-107 | 4.78E-117 | 8.85E-104 | 2.52E-115 | 6.10E-111 |
| 500 | 3.82E-90 | 1.18E-45 | 1.08E-129 | 9.22E-60 | 8.95E-140 | 3.38E-34 | 2.40E-88 | 3.20E-125 | 5.67E-94 | 1.06E-123 | 2.06E-95 |
| 600 | 2.54E-96 | | 2.72E-125 | | 1.70E-153 | | | 4.74E-132 | | 5.63E-132 | |
| 700 | 8.01E-15 | | 1.21E-117 | | 2.61E-151 | | | 2.23E-121 | | 2.26E-121 | |
| 800 | | | 2.86E-110 | | 4.61E-149 | | | 1.92E-107 | | 3.86E-109 | |
| 900 | | | 4.54E-116 | | 2.56E-166 | | | 5.20E-104 | | 1.23E-106 | |
| 1000 | | | 1.32E-105 | | 1.96E-168 | | | 3.61E-111 | | 4.89E-109 | |
| 1100 | | | 1.71E-104 | | 7.24E-175 | | | 1.83E-112 | | 8.40E-112 | |
| 1200 | | | 4.11E-114 | | 8.57E-181 | | | 8.32E-109 | | 1.30E-113 | |
| 1300 | | | 1.67E-85 | | 5.92E-207 | | | 5.41E-80 | | 1.06E-84 | |
| 1400 | | | 7.55E-59 | | 2.71E-191 | | | 3.87E-54 | | 5.70E-65 | |
| 1500 | | | | | 2.44E-169 | | | | | | |
| 1600 | | | | | 7.79E-157 | | | | | | |
| 1700 | | | | | 1.05E-188 | | | | | | |
| 1800 | | | | | 5.17E-199 | | | | | | |
| 1900 | | | | | 4.00E-194 | | | | | | |
| 2000 | | | | | 4.11E-191 | | | | | | |
| 2100 | | | | | 3.57E-210 | | | | | | |